\documentclass[runningheads]{llncs}
\usepackage{graphicx}
\usepackage{comment}
\usepackage{amsmath,amssymb} 
\usepackage{color}

\def\etal{et~al.}			  
\def\ie{i.e.,}               

\newlength\paramargin
\newlength\figmargin
\newlength\secmargin
\newlength\figcapmargin

\newcommand{\Paragraph}[1]
{\vspace{1mm} \noindent\textbf{#1}}

\newcommand{\secref}[1]{Section~\ref{sec:#1}}
\newcommand{\figref}[1]{Figure~\ref{fig:#1}} 
\newcommand{\tabref}[1]{Table~\ref{tab:#1}}

\newcommand{\algref}[1]{Algorithm~\ref{alg:#1}}

\usepackage{booktabs}  
\usepackage{multirow}
\usepackage{makecell}
\usepackage{paralist}
\usepackage{enumitem}

\usepackage[lined,ruled,linesnumbered]{algorithm2e}

\usepackage[width=122mm,left=12mm,paperwidth=146mm,height=193mm,top=12mm,paperheight=217mm]{geometry}
\usepackage[pagebackref=true,breaklinks=true,letterpaper=true,colorlinks,bookmarks=false,citecolor=blue,linkcolor=blue]{hyperref}

\begin{document}
\pagestyle{headings}
\mainmatter
\def\ECCVSubNumber{2943}  

\title{Modeling Artistic Workflows for Image Generation and Editing} 

\titlerunning{Modeling Artistic Workflows for Image Generation and Editing}
%
\author{Hung-Yu Tseng\thanks{Work done during HY's internship at Adobe Research.}$^1$, Matthew Fisher$^2$, Jingwan Lu$^2$, Yijun Li$^2$, Vladimir Kim$^2$, Ming-Hsuan Yang$^{1}$}
\authorrunning{H.-Y. Tseng et al.}
%
\institute{$^1$University of California, Merced\hspace{16pt}$^2$Adobe Research}

\maketitle
\begin{figure}[th]
    \centering
    \includegraphics[width=\linewidth]{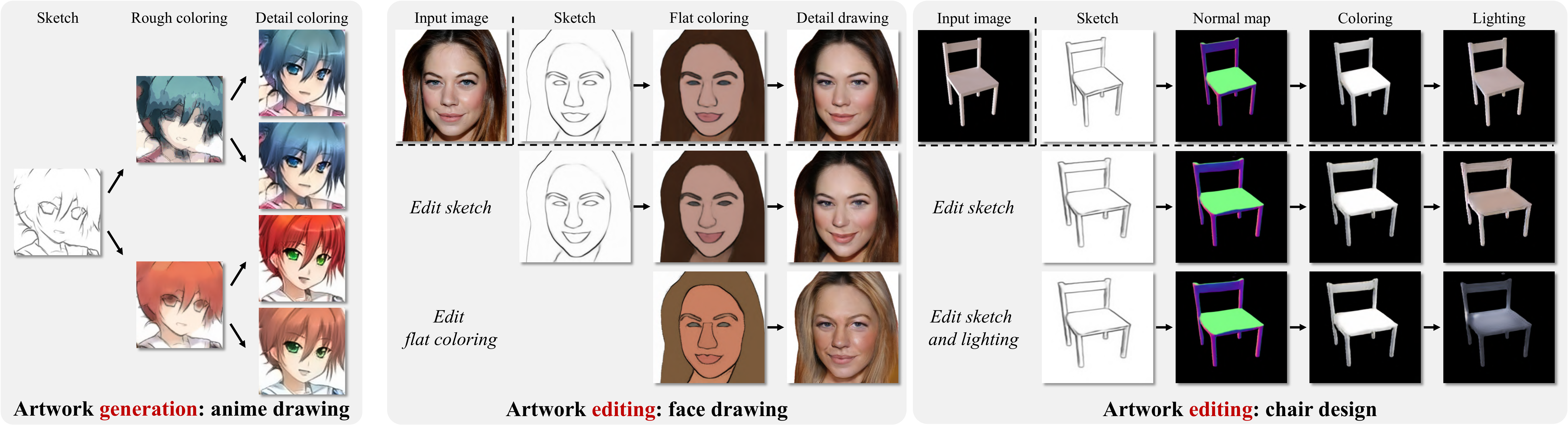}
    \caption{
    We model the sequential creation stages for a given artistic workflow by learning from examples.
    At test time, our framework can guide the user to create new artwork by sampling different variations at each stage (left), and infer the creation stages of existing artwork to enable the user to perform natural edits by exploring variations at different stages (middle and right).}
    \label{fig:teaser}
\end{figure}

\begin{abstract}

People often create art by following an artistic workflow involving multiple stages that inform the overall design.
If an artist wishes to modify an earlier decision, significant work may be required to propagate this new decision forward to the final artwork.
Motivated by the above observations, we propose a generative model that follows a given artistic workflow, enabling both multi-stage image generation as well as multi-stage image editing of an existing piece of art.
Furthermore, for the editing scenario, we introduce an optimization process along with learning-based regularization to ensure the edited image produced by the model closely aligns with the originally provided image.
Qualitative and quantitative results on three different artistic datasets demonstrate the effectiveness of the proposed framework on both image generation and editing tasks.

\end{abstract}

\vspace{-3mm}
\section{Introduction}
\vspace{-1mm}
\label{sec:intro}

Creating artwork from scratch is a herculean task for people without years of artistic experience. 
For novices to the world of art, it would be more feasible to accomplish this task if there are clear creation steps to follow.
Take a watercolor painting for example.
One may be guided to first sketch the outline with pencils, then fill out areas with large brushes, and finalize details such as the color gradient and shadow with small brushes.
At each stage, some aspects~(\ie~variations) of the overall design are determined to carry forward to the final piece of art.

Inspired by these observations, we aim to model workflows for creating art, targeting two relevant artistic applications: multi-stage artwork creation and multi-stage artwork editing.
As shown in~\figref{teaser}, multi-stage artwork generation guides the user through the creation process by starting from the first stage then selecting the variation at each subsequent creation stage.
In the multi-stage artwork editing, we are given a final piece of artwork and infer all the intermediate creation stages, enabling the user to perform different types of editing on various stages and propagate them forward to modify the final artwork.

Existing artwork creation approaches use conditional generative adversarial networks~(conditional GANs)~\cite{isola2017pix2pix,zhu2017toward,DRIT} to produce the artwork according to user-provided input signals.
These methods can take user inputs such as a sketch image~\cite{chen2018sketchygan} or segmentation mask~\cite{park2019SPADE,wang2018pix2pixHD} and perform a single-step generation to synthesize the final artwork.
To make the creation process more tractable, recent frameworks adopt a multi-step generation strategy to accomplish the generation tasks such as fashion simulation~\cite{song2019unsupervised} and sketch-to-image~\cite{ghosh2019interactive}.
However, these approaches typically do not support editing existing artwork.
To manipulate an existing artwork image without degrading the quality, numerous editing schemes~\cite{bau2019semantic,li2018closed,portenier2018faceshop,zhang2017real,zhu2016generative} have been proposed in the past decade.
Nevertheless, these methods either are designed for specific applications~\cite{li2018closed,portenier2018faceshop,zhang2017real} or lack flexible controls over the editing procedure because of the single-stage generation strategy~\cite{bau2019semantic,zhu2016generative}.

In this paper, we develop a conditional GAN-based framework that 1) synthesizes novel artwork via multiple creation stages, and 2) edits existing artwork at various creation stages.
Our approach consists of an artwork generation module and a workflow inference module.
The artwork generation module learns to emulate each artistic stage by a series of multi-modal~(\ie~one-to-many) conditional GAN~\cite{zhu2017toward} networks.
Each network in the artwork generation module uses a stage-specific latent representation to encode the variation presented at the corresponding creation stage.
At test time, the user can determine the latent representation at each stage sequentially for the artwork generation module to synthesize the desired artwork image.

To enable editing existing artwork, we also design an inference module that learns to sequentially infer the corresponding images at all intermediate stages.
We assume a one-to-one mapping from the final to intermediate stages, and use a series of uni-modal conditional GANs~\cite{isola2017pix2pix} to perform this inference.
At test time, we predict the stage-specific latent representations from the inferred images at all intermediate stages.
Depending on the desired type of edit, the user can edit any stage to manipulate the stage-specific image or latent representation and regenerate the final artwork from the manipulated representations.

We observe that directly applying our workflow inference module can cause the reconstructed image to differ slightly from the initially provided artwork.
Such a reconstruction problem is undesirable since the user expects the generated image to be unchanged when no edits are performed.
To address this problem, we design an optimization procedure along with learning-based regularization to refine the reconstructed image.
This optimization aims to minimize the appearance difference between the reconstructed and the original artwork image, while the learning-based regularization seeks to guide the optimization process and alleviate overfitting.

We collect three datasets with different creation stages to demonstrate the use cases of our approach: face drawing, anime drawing, and chair design.
We demonstrate the creation process guided by the proposed framework and present editing results made by artists.
For quantitative evaluations, we measure the reconstruction error and Fr\'echet inception distance~(FID)~\cite{heusel2017gans} to validate the effectiveness of the proposed optimization and learning-based regularization scheme.
We make the code and datasets public available to stimulate the future research.\footnote{\url{https://github.com/hytseng0509/ArtEditing}}

In this work, we make the following three contributions:
\begin{compactitem}
\item We propose an image generation and editing framework which models the creation workflow for a particular type of artwork.
\item We design an optimization process and a learning-based regularization function for the reconstruction problem in the editing scenario.
\item We collect three different datasets containing various design stages and use them to evaluate the proposed approach.
\end{compactitem}
\vspace{-1mm}
\section{Related Work}
\vspace{-1mm}
\label{sec:related}

\Paragraph{Generative adversarial networks~(GANs).}
GANs~\cite{arjovsky2017wasserstein,brock2018large,goodfellow2014generative,karras2017progressive,karras2019style} model the real image distribution via adversarial learning schemes.
Typically, these methods encode the distribution of real images into a latent space by learning the mapping from latent representations to generated images.
To make the latent representation more interpretable, the InfoGAN~\cite{chen2016infogan} approach learns to disentangle the latent representations by maximizing the mutual information.
Similar to the FineGAN~\cite{singh2018finegan} and VON~\cite{zhu2018visual} methods, our approach learns to synthesize an image via multiple stages of generation, and encode different types of variation into separate latent spaces at various stages.
Our framework extends these approaches to also enables image editing of different types of artwork.

\Paragraph{Conditional GANs.}
Conditional GANs learn to synthesize the output image by referencing the input context such as text descriptions~\cite{zhang2018stackgan++}, scene graphs~\cite{tseng2020retrievegan}, segmentation masks~\cite{huang2020semantic,park2019SPADE}, and images~\cite{isola2017pix2pix}.
According to the type of mapping from the input context to the output image, conditional GANs can be categorized as uni-modal~(one-to-one)~\cite{isola2017pix2pix,zhu2017unpaired} or multi-modal~(one-to-many)~\cite{huang2018multimodal,lee2019drit++,MSGAN,zhu2017toward}.
Since we assume there are many possible variations involved for the generation at each stage of the artwork creation workflow, we use the multi-modal conditional GANs to synthesize the next-stage image, and utilize the uni-modal conditional GANs to inference the prior-stage image.

\Paragraph{Image editing.}
Image editing frameworks enable user-guided manipulation without degrading the realism of the edited images.
Recently, deep-learning-based approaches have made significant progress on various image editing tasks such as colorization~\cite{iizuka2016let,larsson2016learning,zhang2016colorful,zhang2017real}, image stylization~\cite{huang2017adain,li2018closed}, image blending~\cite{hung2018learning}, image inpainting~\cite{nazeri2019edgeconnect,pathak2016context}, layout editing~\cite{lee2019neural}, and face editing~\cite{chang2018pairedcyclegan,cheng2020rsegvae,portenier2018faceshop}.
Unlike these task-specific methods, the task-agnostic iGAN~\cite{zhu2016generative} and GANPaint~\cite{bau2019semantic} models map the variation in the training data onto a low-dimensional latent space using GAN models.
Editing can be conducted by manipulating the representation in the learned latent space.
Different from iGAN and GANPaint, we develop a multi-stage generation method to model different types of variation at various stages.

\Paragraph{Optimization for reconstruction.}
In order to embed an existing image to the latent space learned by a GAN model, numerous approaches~\cite{donahue2019bigbigan,larsen2015vaegan,zhu2017toward} propose to train an encoder to learn the mapping from images to latent representations.
However, the generator sometimes fails to reconstruct the original image from the embedded representations.
To address this problem, optimization-based methods are proposed in recent studies.
Abdal~\etal~\cite{abdal2019image2stylegan} and Bau~\etal~\cite{bau2019semantic} adopt the gradient descent scheme to optimize the latent representations and modulations for the feature activations, respectively.
The goal is to minimize the appearance distance between the generated and original images.
We also utilize the optimization strategy to reconstruct existing artwork images.
In addition, we introduce a learning-based regularization function to guide the optimization process.

\Paragraph{Regularizations for deep learning.}
These approaches~\cite{ghiasi2018dropblock,krogh1992simple,larsson2016fractalnet,srivastava2014dropout,dropgrad,crossdomainfewshot} aim to prevent the learning function from overfitting to a specific solution.
Particularly, the weight decay scheme~\cite{krogh1992simple} regularizes by constraining the magnitude of learning parameters during the training phase.
Nevertheless, regularization methods typically involve hyper-parameters that require meticulous hand-tuning to ensure the effectiveness.
The MetaReg~\cite{balaji2018metareg} method designs a learning-to-learn algorithm to automatically find the hyper-parameters of the weight decay regularization to address the domain generalization problem.
Our proposed learning-based regularization is trained with a similar strategy but different objectives to alleviate the overfitting problem described in Section~\ref{sec:3_2}.
\vspace{-1mm}
\section{Method}
\vspace{-1mm}
\label{sec:method}

Our approach is motivated by the sequential creation stages of artistic workflows.
We build a model that enables a user to 1) follow the creation stages to generate novel artwork and 2) conduct edits at different stages.
Our framework is composed of an artwork generation and a workflow inference module. 
As shown in~\figref{framework}(a), the artwork generation module learns to model the creations stages of the artist workflow.
To enable editing an existing piece of art, the workflow inference module is trained to sequentially infer the corresponding images at all creation stages.
When editing existing artwork, it is important that the artwork remains as close as possible to the original artwork, and only desired design decisions are altered. To enable this, we design an optimization process together with a learning-based regularization that allows faithful reconstruction of the input image.
We provide the implementation and training details for each component in the proposed framework as supplemental material.

\subsection{Artwork Generation and Workflow Inference}
\vspace{-1mm}
\label{sec:3_1}

\begin{figure}[t]
\centering
\includegraphics[width=0.98\linewidth]{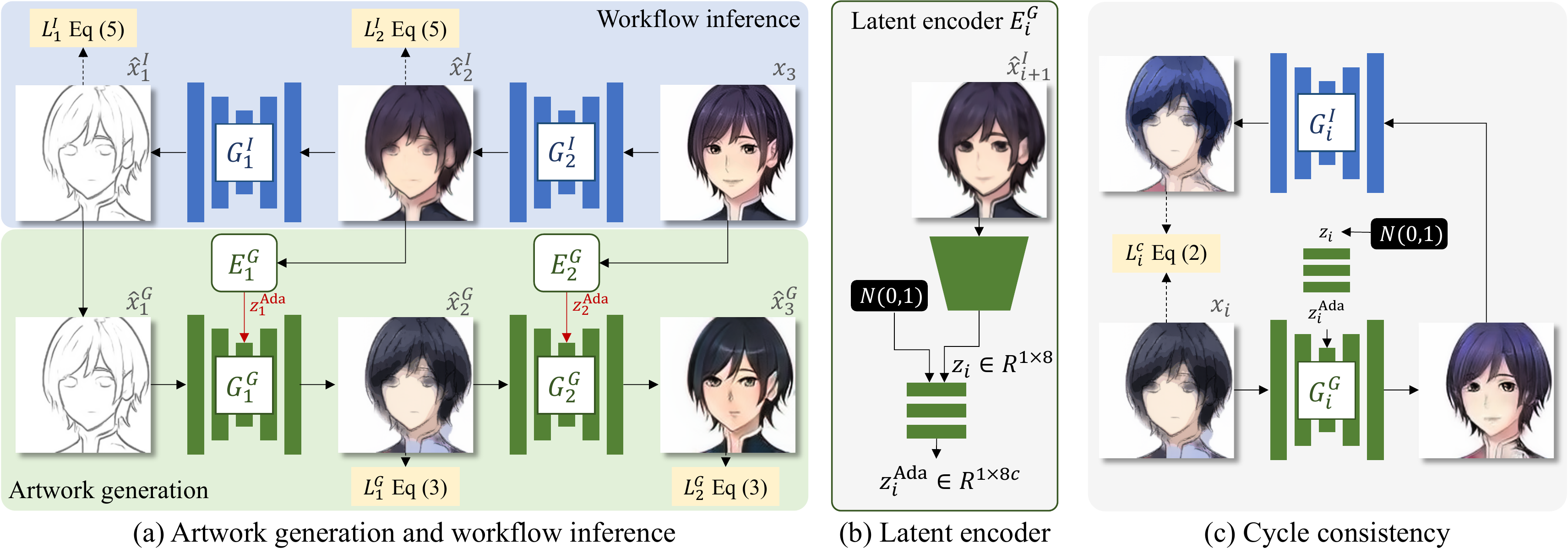}
\caption{\textbf{Overview of the proposed framework}. (a) Given $N$ creation stages ($N=3$ in this example), our approach consists of $N-1$ workflow inference networks and $N-1$ artwork generation networks. The workflow inference module produces the intermediate results of the input artwork at all creation stages. The artwork generation module computes the latent representation $z$ and transformation parameter $z^\mathrm{Ada}$ for each stage, then reconstructs the input artwork images from these transformation parameters. (b) The latent encoder $E_i^G$ extracts the stage-specific latent representation $z$ from the example, and computes the transformation parameters $z^\mathrm{Ada}$ for the AdaIN normalization layers ($c$ channels). (c) We introduce a cycle consistency loss for each stage to prevent the artwork generation model~(which accounts for detail coloring in this example) from memorizing the variation determined at the previous stages~(sketching and flat coloring).}
\vspace{-3mm}
\label{fig:framework}
\end{figure}

\Paragraph{Preliminaries.}
The proposed approach is driven by the number of stages in the training dataset and operates in a supervised setting with aligned training data.
Denoting $N$ as the number of stages, the training dataset is comprised of a set of image groups $\{(x_1,x_2,\cdots,x_{N})\}$, where $x_N$ denotes the artwork image at the final stage.
We construct the proposed framework with $N-1$ workflow inference models 
$\{G^I_i\}^N_{i=1}$ as well as $N-1$ artwork generation models $\{(E^G_i,G^G_i)\}^N_{i=1}$.
We show an example of $3$ stages in~\figref{framework}(a).
Since the proposed method is based on the observation that artists sequentially determine a design factor~(\ie~variation) at each stage, we assume that the generation from the image in the prior stage to the later one is multi-modal (\ie~one-to-many mapping), while the inference from the final to the previous stages is uni-modal (\ie~one-to-one mapping).

\Paragraph{Artwork generation.}
The artwork generation module aims to mimic the sequential creation stages of the artistic workflow.
Since we assume the generation from the prior stages to the following ones is multi-modal, we construct a series of artwork generation networks by adopting the multi-modal conditional GAN approach in BicycleGAN~\cite{zhu2017toward} and the network architecture of MUNIT~\cite{huang2018multimodal}.
As shown in~\figref{framework} (a) and (b), each artwork generation model contains two components: latent encoder $E^G_i$ and generator $G^G_i$. 
The latent encoder $E^G_i$ encodes the variation presented at the $i$-th stage in a stage-specific latent space.
Given an input image $x_i$ and the corresponding next-stage image $x_{i+1}$, the latent encoder $E^G_i$ extracts the stage-specific latent representation $z_i$ from the image $x_{i+1}$, and computes the transformation parameter $z^\mathrm{Ada}_i$.
The generator $G^G_i$ then takes the current-stage image $x_i$ as input and modulates the activations through the AdaIN normalization layers~\cite{zhu2017toward} with the transformation parameter $z^\mathrm{Ada}_i$ to synthesize the next-stage image $\hat{x}^G_{i+1}$, namely
\begin{equation}
\hat{x}^G_{i+1}=G^G_i(x_i,E^G_i(x_{i+1}))\hspace{5mm}i\in\{1,2,\cdots,N-1\}.
\end{equation}
We utilize the objective introduced in the BicycleGAN~\cite{zhu2017toward}, denoted as $L^\mathrm{bicycle}_i$, for training the generation model.
The objective $L^\mathrm{bicycle}_i$ is detailed in the supplementary material.

Ideally, the artwork generation networks corresponding to a given stage would encode only new information~(\ie~incremental variation), preserving prior design decisions from earlier stages.
To encourage this property, we impose a cycle consistency loss to enforce the generation network to encode the variation presented at the current stage only, as shown in~\figref{framework}(c).
Specifically, we use the inference model $G^I_i$ to map the generated next-stage image back to the current stage.
The mapped image should be identical to the original image $x_i$ at the current stage, namely
\begin{equation}
L^\mathrm{c}_i=\lVert G^I_i(G^G_i(x_i,E^G_i(z_i)))-x_i\rVert_1\hspace{5mm}z_i\sim N(0,1).
\end{equation}
Therefore, the overall training objective for the artwork generation model at the $i$-th stage is
\begin{equation}
L^G_i=L^\mathrm{bicycle}_i + \lambda^cL^c_i,
\end{equation}
where $\lambda^c$ controls the importance of the cycle consistency.

\Paragraph{Workflow inference.}
To enable the user to edit the input artwork $x_N$ at different creation stages, our inference module aims to hallucinate the corresponding images at all previous stages.
For the $i$-th stage, we use a unimodal conditional GAN network~\cite{isola2017pix2pix} to generate the image at $i$-th stage from the image at ($i+1$)-th stage, namely
\begin{equation}
\hat{x}^I_{i}=G^I_i(x_{i+1})\hspace{5mm}i\in\{1,2,\cdots,N-1\}.
\end{equation}
During the training phase, we apply the hinge version of GAN loss~\cite{brock2018large} to ensure the realism of the generated image $\hat{x}^I_{i}$.
We also impose an $\ell_1$ loss between the synthesized image $\hat{x}^I_{i}$ and the ground-truth image $x_i$ to stabilize and accelerate the training.
Hence the training objective for the inference network at the $i$-th stage is
\begin{equation}
L^I_i=L^\mathrm{GAN}_i(\hat{x}^I_i) + \lambda^\mathrm{1}\lVert\hat{x}^I_{i}-x_i\rVert_1,
\end{equation}
where $\lambda^\mathrm{1}$ controls the importance of the $\ell_1$ loss.

\Paragraph{Test-time inference.} As shown in~\figref{framework}(a), given an input artwork image $x_N$, we sequentially obtain the images at all previous stages $\{\hat{x}^I_i\}^N_{i=1}$ using the workflow inference module~(blue block).
We then use the artwork generation module~(green block) to extract the latent representations $\{z_i\}^{N-1}_{i=1}$ from the inferred images $\{\hat{x}^I_i\}^N_{i=1}$, and compute the transformation parameters $\{z^\mathrm{Ada}_i\}^{N-1}_{i=1}$.
Combining the first-stage image $x^G_1=x^I_1$ and the transformation parameters $\{z^\mathrm{Ada}_i\}^{N-1}_{i=1}$, the generation module consecutively generates the images $\{\hat{x}^G_i\}^N_{i=2}$ at the following stages.
The user can choose the stage to manipulate based on the type of edit desired.
Edits at the $i$-th stage can be performed by either manipulating the latent representation $z_i$ or directly modifying the image $x^G_i$.
For example, in~\figref{framework}(a), the user can choose to augment the representation $z_1$ to adjust the flat coloring.
After editing, the generation module generates the new artwork image at the final stage.

\vspace{-1mm}
\subsection{Optimization for Reconstruction}
\vspace{-1mm}
\label{sec:3_2}

\begin{figure}[t]
\centering
\includegraphics[width=0.65\linewidth]{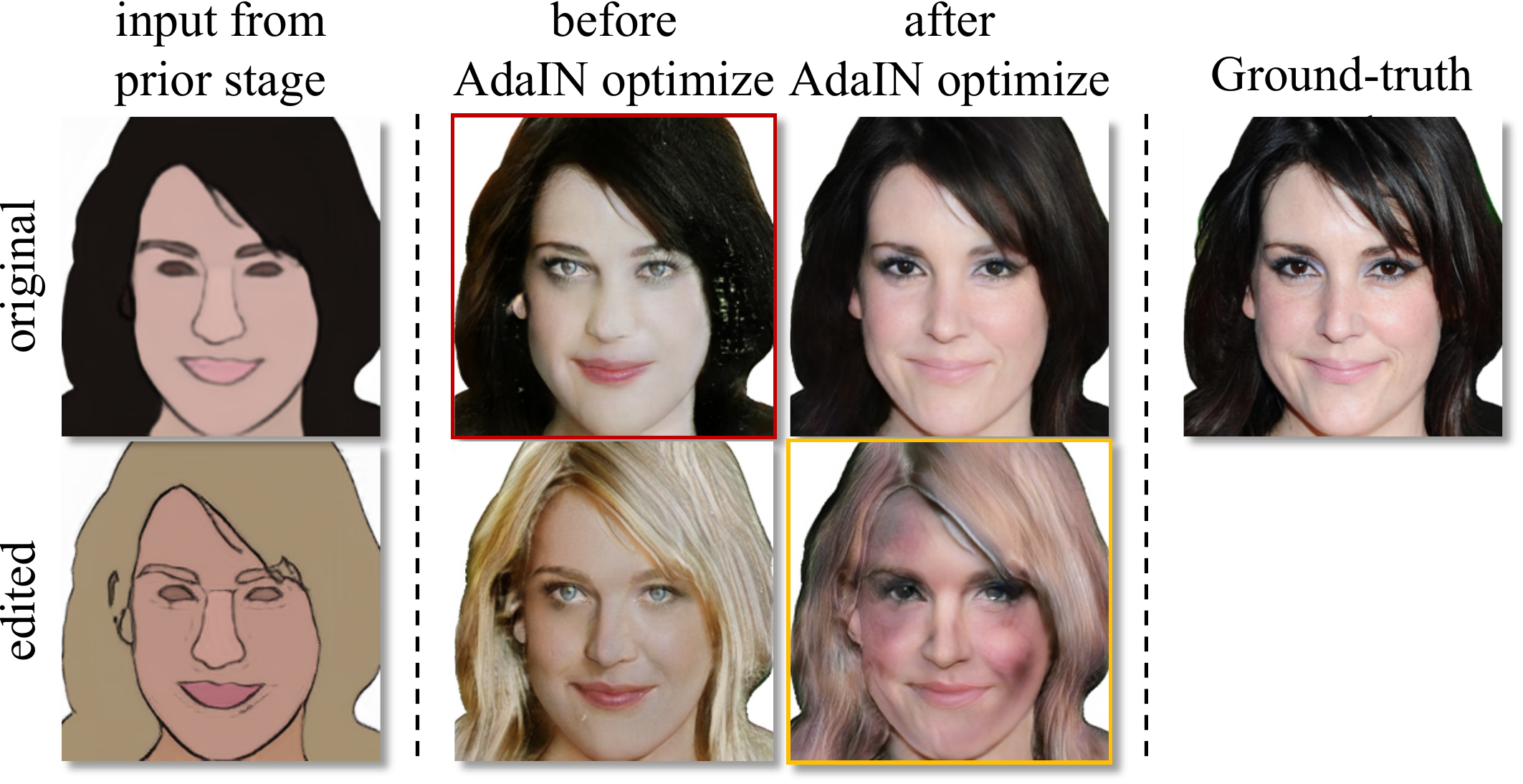}
\vspace{-3mm}
\caption{\textbf{Motivation of the AdaIN optimization and learning-based regularization.} The proposed AdaIN optimization and the learning-based regularization are motivated by the observations that 1) using the computed transformation parameters $z^\mathrm{Ada}$ in~\figref{framework} cannot well reconstruct the original input image~(red outline in 1-st row)), and 2) the AdaIN optimization may degrade the quality of the editing results~(yellow outline in 2-nd row).}
\vspace{-3mm}
\label{fig:Ada}
\end{figure}

As illustrated in~\secref{3_1}, the artwork generation module would ideally reconstruct the input artwork image~(\ie~$\hat{x}^G_N=x_N$) from the transformation parameters $\{z^\mathrm{Ada}_i\}^{N-1}_{i=1}$ before the user performs an edit.
However, the reconstructed image $\hat{x}^G_N$ may be slightly different from the input image $x_N$, as shown in the first row of~\figref{Ada}.
Therefore, we adopt an AdaIN optimization algorithm to optimize the transformation parameters $\{z^\mathrm{Ada}_i\}^N_{i=1}$ of the AdaIN normalization layers in the artwork generation models.
The goal of the AdaIN optimization is to minimize the appearance distance between the reconstructed and input image.

While this does improve the reconstruction of the input image, we observe that the optimization procedure causes the generation module to memorize input image details, which degrades the quality of some edited results, as shown in the second row of~\figref{Ada}.
To mitigate this memorization, we propose a learning-based regularization to improve the AdaIN optimization.

\Paragraph{AdaIN optimization.}
The AdaIN optimization approach aims to minimize the appearance distance between the reconstructed image $\hat{x}^G_N$ and the input artwork image $x_N$.
There are many choices for what to optimize to improve reconstruction: we could optimize the parameters in the generation models or the extracted representations $\{z_i\}^N_{i=1}$. 
Optimizing model parameters is inefficient because of the large number of parameters to be updated.
On the other hand, we find that optimizing the extracted representation is ineffective, as validated in~\secref{quantitative}.
As a result, we choose to optimize the transformation parameters $\{z^\mathrm{Ada}_i\}^N_{i=1}$ of the AdaIN normalization layers in the generation models, namely the AdaIN optimization.
Note that a recent study~\cite{abdal2019image2stylegan} also adopts a similar strategy.

We conduct the AdaIN optimization for each stage sequentially.
The transformation parameter at the early stage is optimized and then fixed for the optimization at the later stages.
Except for the last stage~(\ie~$i=N-1$) that uses the input artwork image $x_N$, the inferred image $x^I_{i+1}$ by the inference model serves as the reference image $x^\mathrm{ref}$ for the optimization.
For each stage, we first use the latent encoder $E^G_i$ to compute the transformation parameter $z^\mathrm{Ada}_i$ from the reference image for generating the image.
Since there are four AdaIN normalization layers with $c$ channels in each artwork generation model, the dimension of the transformation parameter is $1\times{8c}$ (a scale and a bias term for each channel).
Then we follow the standard gradient descent procedure to optimize the transformation parameters with the goal of minimizing the loss function $L^\mathrm{Ada}$ which measures the appearance distance between the synthesized image $\hat{x}^G_i$ by the generator $G^G_i$ and the reference image $x^\mathrm{ref}$.
The loss function $L^\mathrm{Ada}$ is a combination of the pixel-wise $\ell_1$ loss and VGG-16 perceptual loss~\cite{johnson2016perceptual}, namely
\begin{equation}
L^\mathrm{Ada}(\hat{x}^G_i,x^\mathrm{ref}) = \lVert \hat{x}^G_i - x^\mathrm{ref} \rVert_1 + \lambda^pL^p(\hat{x}^G_i,x^\mathrm{ref}),
\label{eq:ada}
\end{equation}
where $\lambda_p$ is the importance term.
We summarize the AdaIN optimization in~\algref{AdaIN}.
Note that in practice, we optimize the incremental term $\delta^\mathrm{Ada}_i$ for the transformation parameter $z^\mathrm{Ada}_i$, instead of updating the parameter itself.

\begin{figure}[t]
\centering
\includegraphics[width=0.65\linewidth]{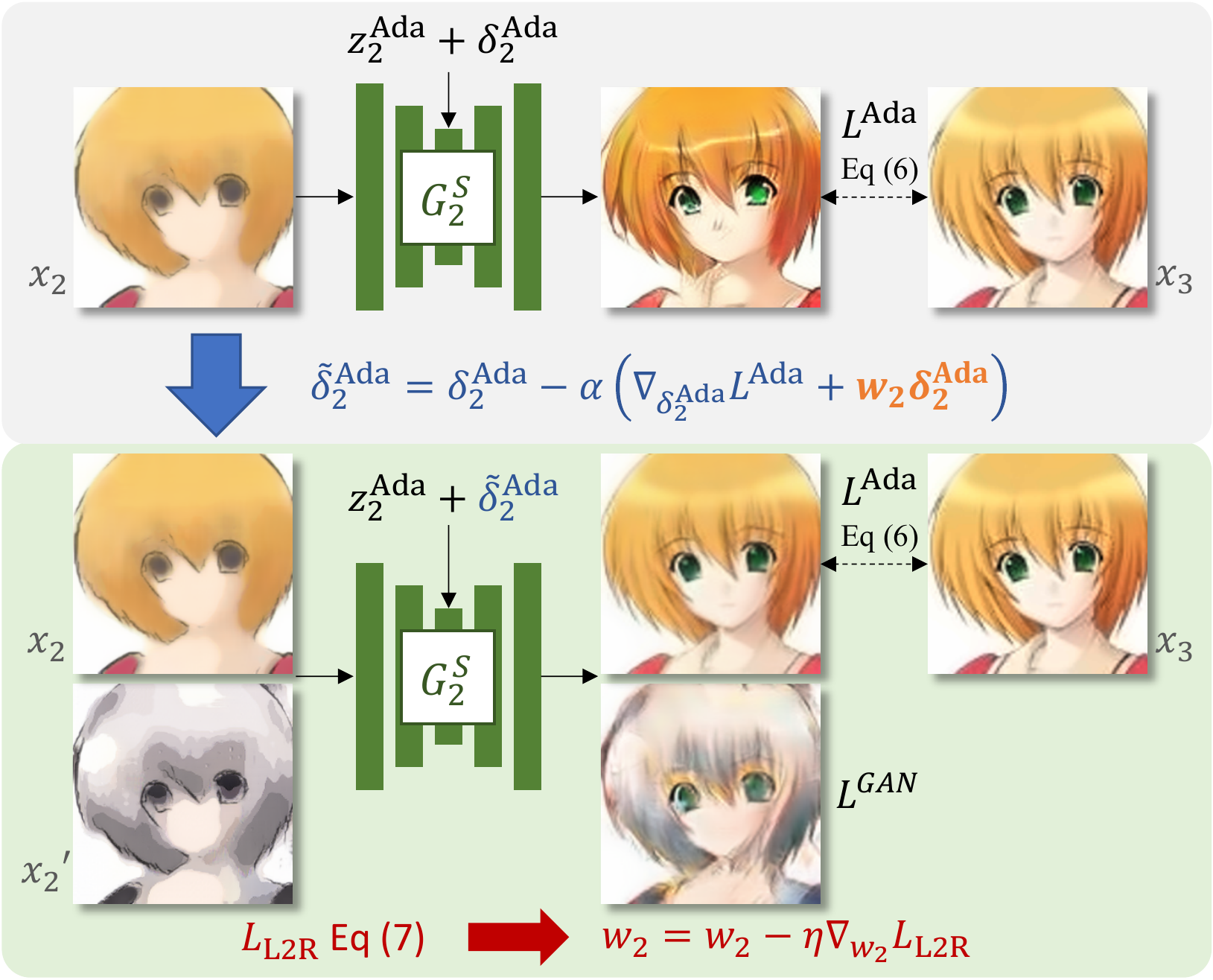}
\vspace{-2mm}
\caption{\textbf{Training process for learning-based regularization.} For the $i$-th stage ($i=2$ in this example), we optimize the hyper-parameter $w_i$ for the weight decay regularization~(orange text) by involving the AdaIN optimization in the training process: after the incremental term $\delta^\mathrm{Ada}_i$ is updated via one step of AdaIN optimization and the weight decay regularization~(blue arrow), the generation model should achieve improved reconstruction as well as maintain the quality of the editing result~(green block). Therefore, we use the losses $L^\mathrm{Ada},L^\mathrm{GAN}$ computed from the updated parameter $\Tilde{\delta}^\mathrm{Ada}_i$ to optimize the hyper-parameter $w_i$~(red arrow).}
\vspace{-5mm}
\label{fig:l2r}
\end{figure}

\Paragraph{Learning-based regularization.}
Although the AdaIN optimization scheme addresses the reconstruction problem, it often degrades the quality of editing operations, as shown in the second row of~\figref{Ada}.
This is because the AdaIN optimization causes overfitting (memorization of the reference image $x^\mathrm{ref}$).
The incremental term $\delta^\mathrm{Ada}_i$ for the transformation parameter $z^\mathrm{Ada}_i$ is updated to extreme values to achieve better reconstruction, so the generator becomes sensitive to the change~(\ie~editing) on the input image and produces unrealistic results.

To address the overfitting problem, we use weight decay regularization~\cite{krogh1992simple} to constrain the magnitude of the incremental term $\delta^\mathrm{Ada}_i$, as shown in Line 6 in~\algref{AdaIN}.
However, it is difficult to find a general hyper-parameter setting $w_i\in R^{1\times{8c}}$ for different generation stages of various artistic workflows.
Therefore, we propose a learning algorithm to optimize the hyper-parameter $w_i$.
The core idea is that updating the incremental term $\delta^\mathrm{Ada}_i$ with the regularization $w_i\delta^\mathrm{Ada}_i$ should 1) improve the reconstruction and 2) maintain the realism of edits on an input image.
We illustrate the proposed algorithm in~\figref{l2r}.
In each iteration of training at the $i$-th stage, we sample an image pair $(x_i, x_{i+1})$ and an additional input image $x'_i$ from the training dataset.
The image $x'_i$ serves as the edited image of $x_i$.
We first use the latent encoder $E^G_i$ to extract the transformation parameter $z^\mathrm{Ada}_i$ from the next-stage image $x_{i+1}$.
As shown in the grey block of~\figref{l2r}, we then update the incremental term from $\delta^\mathrm{Ada}_i$ to $\Tilde{\delta}^\mathrm{Ada}_i$ via one step of the AdaIN optimization and the weight decay regularization.
With the updated incremental term $\Tilde{\delta}^\mathrm{Ada}_i$, we use the loss function $L^\mathrm{Ada}$ to measure the reconstruction quality, and use the GAN loss to evaluate the realism of editing results, namely
\begin{equation}
\begin{aligned}
L^\mathrm{L2R}&=L^\mathrm{Ada}(G^G_i(x_i,z^\mathrm{Ada}_i+\Tilde{\delta}^\mathrm{Ada}_i), x_{i+1})\\
&+\lambda^\mathrm{GAN}L^\mathrm{GAN}(G^G_i(x'_i,z^\mathrm{Ada}_i+\Tilde{\delta}^\mathrm{Ada}_i)).
\end{aligned}
\end{equation}
Finally, since the loss $L^\mathrm{L2R}$ indicates the efficacy of the weight decay regularization, we optimize the hyper-parameter $w_i$ by
\begin{equation}
w_i=w_i - \eta\bigtriangledown_{w_i}L^\mathrm{L2R},
\end{equation}
where $\eta$ is the learning rate of the training algorithm for the proposed learning-based regularization.

\begin{algorithm}[t]
  \caption{AdaIN optimization at $i$-th stage}
  \label{alg:AdaIN}
  \DontPrintSemicolon
  \textbf{Require:} reference image $x^\mathrm{ref}=x_N$ or $x^\mathrm{ref}=\hat{x}^I_{i+1}$, input image $\hat{x}^G_{i}$, learning rate $\alpha$, iterations $T$,  regularization parameter $w_i$\;
  \vspace{1mm}
  $z^\mathrm{Ada}_i=E^G_i(x^\mathrm{ref})$,\hspace{2mm} $\delta^\mathrm{Ada}_i=\mathbf{0}\in R^{1\times{8c}}$\;
  \vspace{1mm}
  \While{$t=\{1,\ldots,T\}$} {
    \vspace{1mm}
    $\hat{x}^G_{i+1}=G^G_i(\hat{x}^G_{i},z^\mathrm{Ada}_i+\delta^\mathrm{Ada}_i)$\;
    \vspace{1mm}
    $L^\mathrm{Ada}=\lVert\hat{x}^G_{i+1}-x^\mathrm{ref}\rVert_1 + \lambda^p L^p(\hat{x}^G_{i+1},x^\mathrm{ref})$\;
    \vspace{1mm}
    $\delta^\mathrm{Ada}_i=\delta^\mathrm{Ada}_i-\alpha\left(\bigtriangledown_{\delta^\mathrm{Ada}_i}L_\mathrm{Ada} + w_i\delta^\mathrm{Ada}_i\right)$\;
  }
  \textbf{Return:} $z^\mathrm{Ada}_i+\delta^\mathrm{Ada}_i$\;
\end{algorithm}

\vspace{-1mm}
\section{Experimental Results}
\vspace{-1mm}
\label{sec:results}

\subsection{Datasets}
\vspace{-1mm}
To evaluate our framework, we manually process face drawing, anime drawing, and chair design datasets.
\tabref{dataset} summarizes the generation stages, the number of training images, the number of testing images, and the source of the images for each dataset.
%
%
We describe more details in the supplementary material.

\subsection{Qualitative Evaluation}
\vspace{-1mm}
\begin{table}[t]
    \scriptsize
	\centering
	\caption{\textbf{Summarization of the datasets.} Three datasets are processed for evaluating the proposed framework.}
	\begin{tabular}{@{}l lll@{}} 
	    \toprule
	    Dataset & Face drawing & Anime drawing & Chair design\\
	    \midrule
	    Source & CelebaHQ~\cite{CelebAMask-HQ} & EdgeConnect~\cite{nazeri2019edgeconnect} & ShapeNet~\cite{chang2015shapenet} \\
		\# Training images & 29000 & 33323 & 12546 \\
		\# Testing images & 1000 & 1000 & 1000 \\
		\makecell[l]{Stages} & \makecell[l]{1. sketch\\2. flat coloring\\3. detail drawing} & \makecell[l]{1. sketch\\2. rough coloring\\3. detail coloring} & \makecell[l]{1. sketch\\2. normal map\\3. coloring\\4. lighting}\\
		\bottomrule 
	\end{tabular}
	\label{tab:dataset}
	\vspace{-3mm}
\end{table}

\begin{figure*}[t]
\centering
\includegraphics[width=\linewidth]{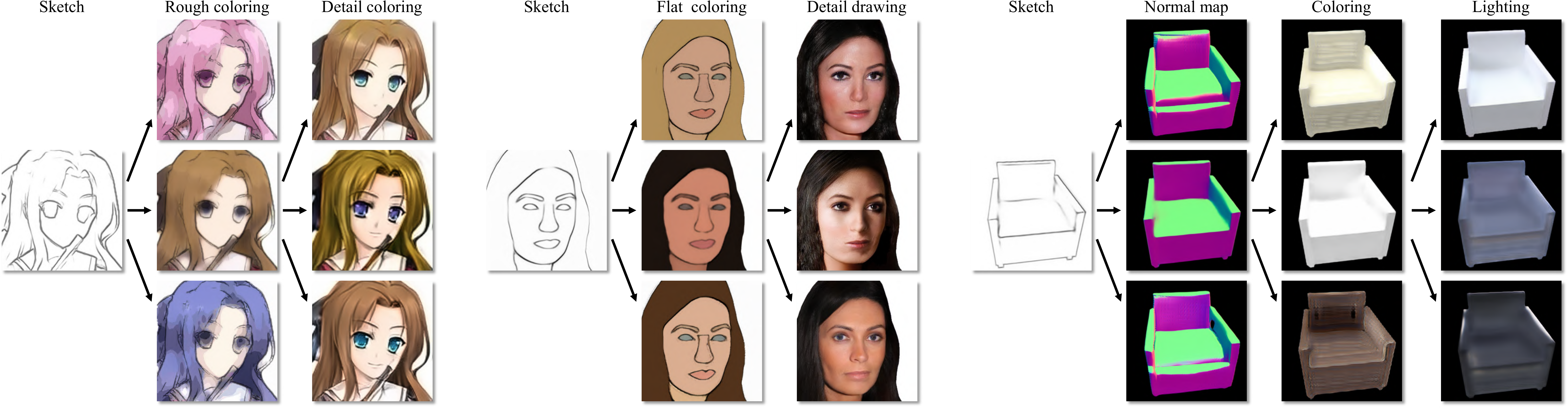}
\caption{\textbf{Results of image generation from the first stage.} We use the first-stage testing images as input and randomly sample the latent representations to generate the image at the final stage.}
\vspace{-3mm}
\label{fig:creation}
\end{figure*}

\Paragraph{Generation.}
We present the generation results at all stages in~\figref{creation}.
In this experiment, we use the testing images at the first stage as inputs, and randomly sample various latent representation $z\in\{z_i\}^{N-1}_{i=1}$ at each stage of the proposed artwork generation module. 
The generation module sequentially synthesizes the final result via multiple stages. It successfully generates variations by sampling different random latent codes at different stages.
For example, when generating anime drawings, manipulating the latent code at the final stage produces detailed color variations, such as modifying the saturation or adding the highlights to the hair regions.

\begin{figure*}[t]
\centering
\includegraphics[width=\linewidth]{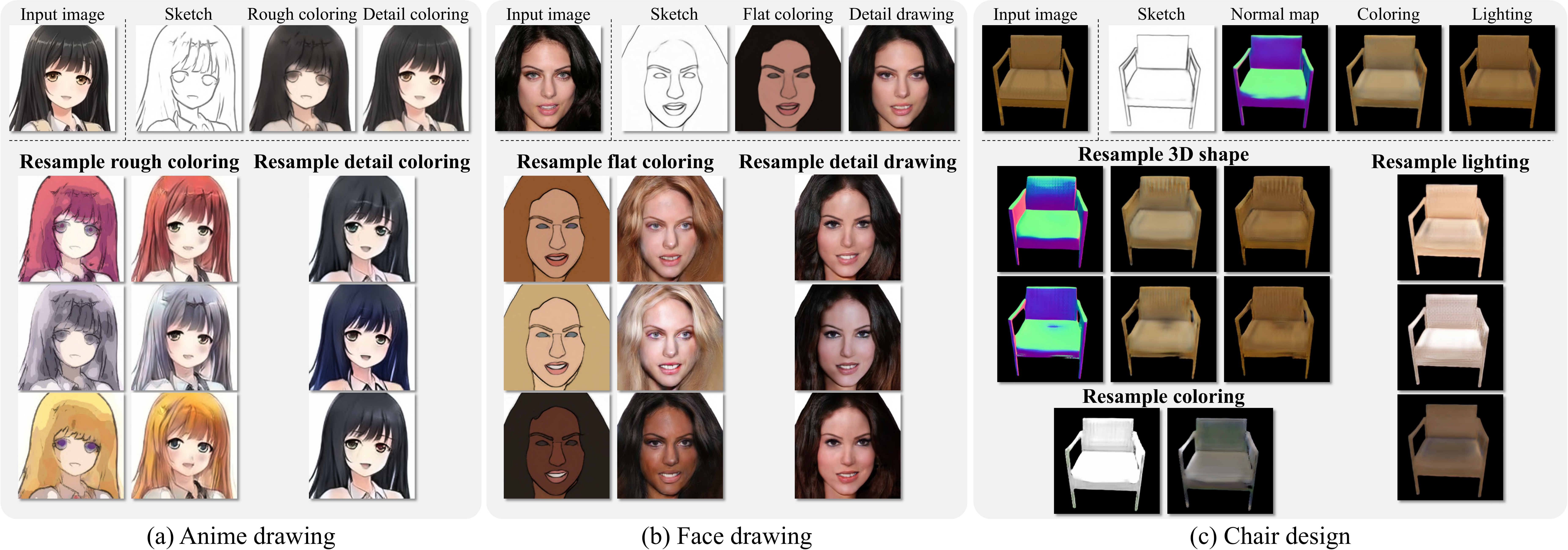}
\caption{\textbf{Re-sampling latent representation at each stage.} After we use the AdaIN optimization process to reconstruct the input image~(1st row), we edit the reconstructed image by re-sampling the latent representations at various stages.}
\vspace{-3mm}
\label{fig:radomedit}
\end{figure*}

\Paragraph{Editing.}
\figref{radomedit} shows the results of editing the artwork images at different stages.
Specifically, after the AdaIN optimization reconstructs the testing image at the final stage~(first row), we re-sample the representations $z\in\{z_i\}^{N-1}_{i=1}$ at various stages. 
Our framework is capable of synthesizing the final artwork such that its appearance only changes with respect to the stage with re-sampled latent code.
For example, for editing face drawings, re-sampling representations at the flat coloring stage only affects hair color, while maintaining the haircut style and details.

\begin{figure*}[t]
\centering
\includegraphics[width=0.98\linewidth]{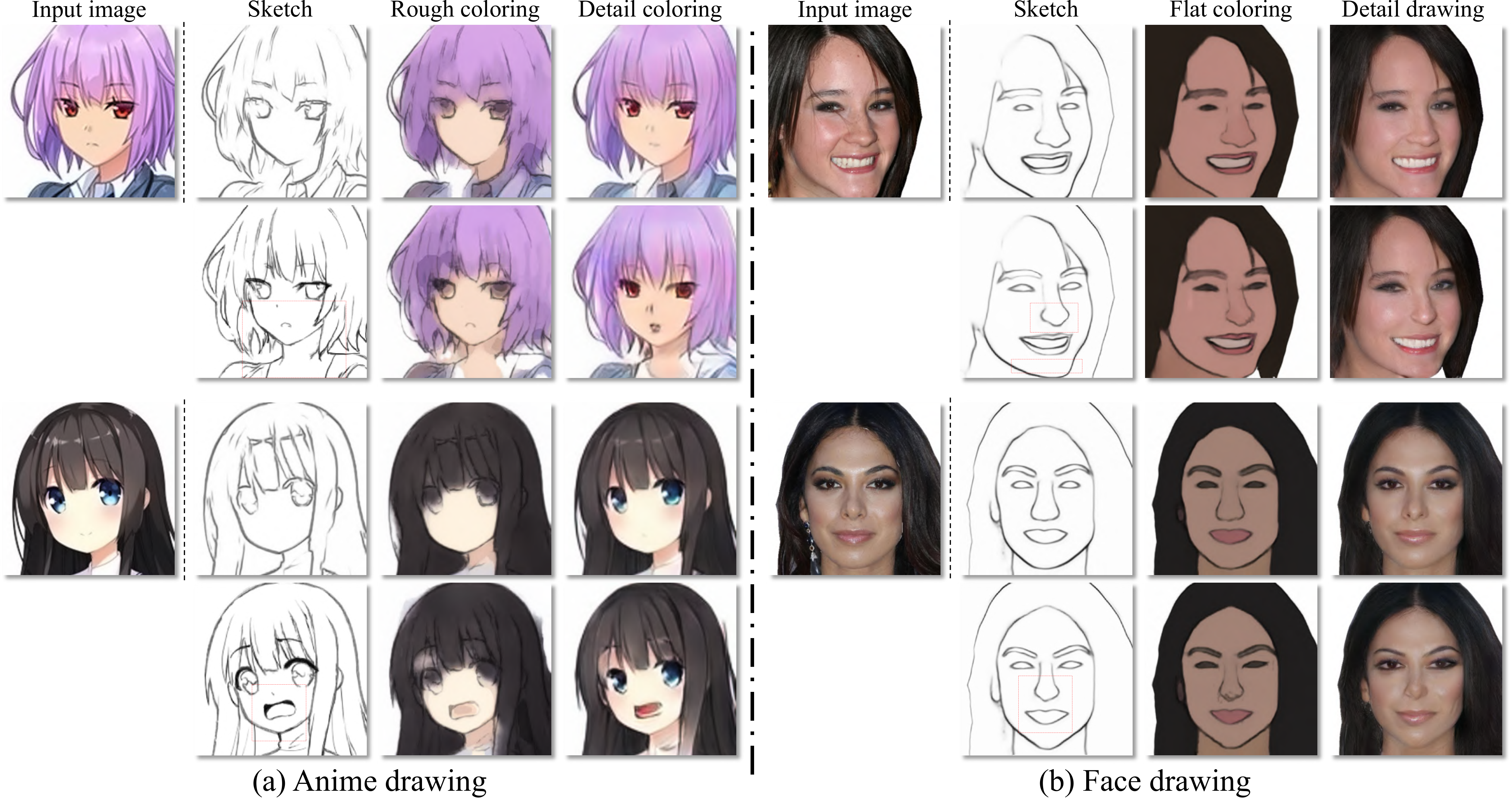}
\caption{\textbf{Results of artistic editing.} Given an input artwork image, we ask the artist to edit the inferred sketch image. The synthesis model then produces the corresponding edited artwork. The first row shows the input artwork and inferred images, and the red outlines indicate the edited regions.}
\vspace{-4mm}
\label{fig:artistedit}
\end{figure*}

To evaluate the interactivity of our system, we also asked professional artists to edit some example sketches (\figref{artistedit}).
First, we use the proposed framework to infer the initial sketch from the input artwork image.
Given the artwork image and the corresponding sketch, we asked an artist to modify the sketch manually.
For the edited sketch (second row), we highlight the edits with the red outlines.
This experiment confirms that the proposed framework enables the artists to adjust only some stages of the workflow, controlling only desired aspects of the final synthesized image. Additional artistic edits are shown in Figure~\ref{fig:teaser}.

\begin{figure}[t]
\centering
\includegraphics[width=0.8\linewidth]{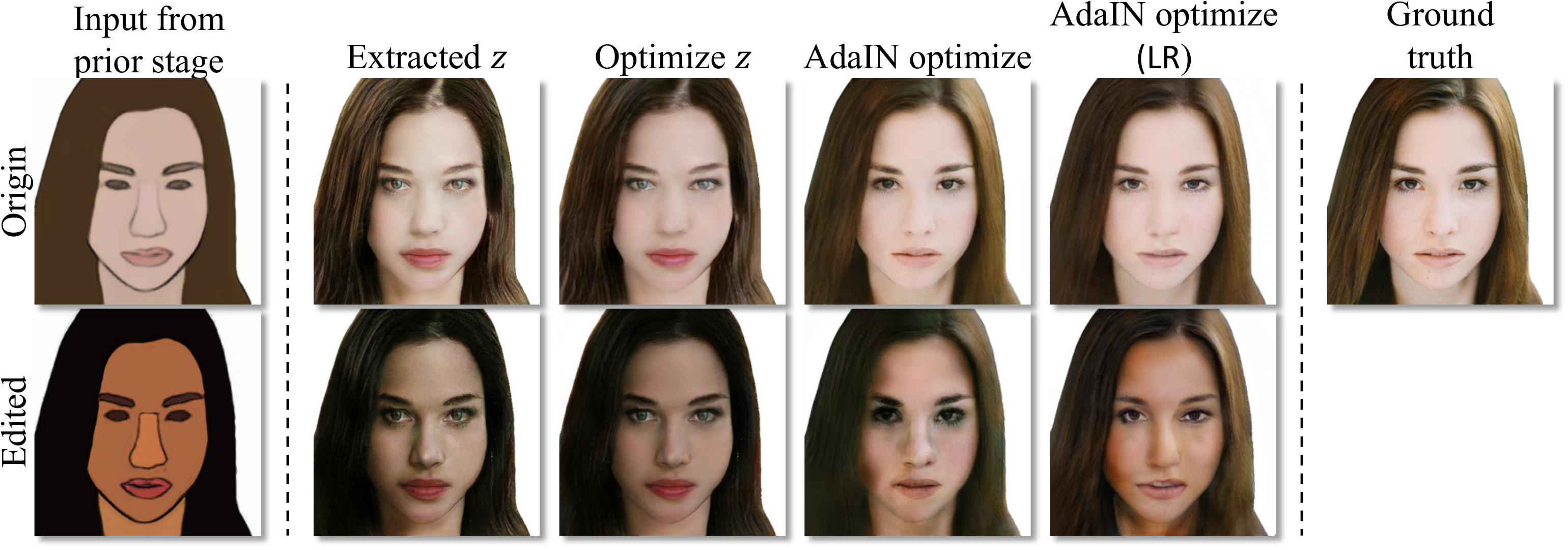}
\caption{\textbf{Results of different optimization approaches.} We show both the reconstruction and editing results of various optimization approaches at the final stage for the face drawing dataset.}
\vspace{-2mm}
\label{fig:adainresult}
\end{figure}

\Paragraph{AdaIN optimization and learning-based regularization.}
\figref{adainresult} presents the results of the AdaIN optimization and the proposed learning-based regularization.
As shown in the first row, optimizing representations $z$ fails to refine the reconstructed images due to the limited capacity of the low-dimensional latent representation.
In contrast, the AdaIN optimization scheme minimizes the perceptual difference between the input and reconstructed images.
We also demonstrate how the optimization process influences the editing results in the second row.
Although the AdaIN optimization resolves the reconstruction problem, it leads to overfitting and results in unrealistic editing results synthesized by the generation model.
By utilizing the proposed learning-based regularization, we address the overfitting problem and improve the quality of the edited images.

\subsection{Quantitative Evaluation}
\vspace{-1mm}
\label{sec:quantitative}
\Paragraph{Evaluation metrics.}
We use the following metrics in the quantitative evaluation.
\begin{compactitem}
\item{Reconstruction error:} Given the input artwork $x_N$ and the reconstructed image $\hat{x}^G_N$, we use the $\ell_1$ distance $\lVert\hat{x}^G_N-x_N\rVert$ to evaluate the reconstruction quality.
\item{FID:} We use the Fr\'echet Inception Distance (FID)~\cite{heusel2017gans} score to measure the realism of generated images $\hat{x}^G_N$. A smaller FID score indicates better visual quality.
\end{compactitem}

\Paragraph{Reconstruction.}
As shown in~\secref{3_2}, we conduct the AdaIN optimization for each stage sequentially to reconstruct the testing image at the final stage. 
We use both the reconstruction error and FID score to evaluate several baseline methods and the AdaIN optimization, and show the results in \tabref{recon}.
Results on the 2-nd and 3-rd rows demonstrate that the AdaIN optimization is more effective than optimizing the latent representations $\{z_i\}^{N-1}_{i=1}$.
On the other hand, applying stronger weight decay regularization~(\ie~$w_i=10^{-2}$) diminishes the reconstruction ability of the AdaIN optimization.
By applying the weight decay regularization with learned hyper-parameter $w$~(\ie~LR), we achieve comparable reconstruction performance in comparison to the optimization without regularization.

\begin{table}[t]
    \footnotesize
	\centering
	\caption{\textbf{Quantitative results of reconstruction.} We use the $\ell1$ pixel-wise distance~($\downarrow$) and the FID~($\downarrow$) score to evaluate the reconstruction ability. $w$ and LR indicates the hyper-parameter for the weight regularization and applying the learned regularization, respectively.}
	\vspace{-2mm}
	\begin{tabular}{@{}ll cccccc} 
	    \toprule
		\multirow{2}{*}{\shortstack[l]{Optimization}} & \multirow{2}{*}{\shortstack[l]{$w$}}& \multicolumn{2}{c}{Face} & \multicolumn{2}{c}{Anime} & \multicolumn{2}{c}{Chair} \\
		\cmidrule(lr){3-8} 
		& & $\ell1$ & FID & $\ell1$ & FID & $\ell1$ & FID \\
		\midrule
		None & - & $0.094$ & $39.78$ & $0.127$ & $36.73$ & $0.074$ & $129.2$ \\
		$z$ & $0$ & $0.104$ & $40.70$ & $0.126$ & $45.66$ & $0.068$ & $107.0$ \\
		AdaIN & $0$ & $\mathbf{0.040}$ & $\mathbf{34.61}$ & $\mathbf{0.042}$ & $\mathbf{26.56}$ & $\mathbf{0.009}$ & $\mathbf{46.48}$\\
		AdaIN & $10^{-3}$ & $0.043$ & $35.78$ & $0.056$ & $29.14$ & $0.019$ & $53.08$\\
		AdaIN & $10^{-2}$ & $0.053$ & $39.19$ & $0.097$ & $46.31$ & $0.049$ & $83.58$\\
		\midrule
		AdaIN & LR & $0.045$ & $33.28$ & $0.070$ & $34.16$ & $0.018$ & $49.44$\\
		\bottomrule 
	\end{tabular}
	\label{tab:recon}
	\vspace{-4mm}
\end{table}

\Paragraph{Editing.}
In this experiment, we investigate how various optimization methods influence the quality of edited images.
For each testing final-stage image, we first use different optimization approaches to refine the reconstructed images.
We then conduct the editing by re-sampling the latent representation $z_i$ at a randomly chosen stage.
We adopt the FID score to measure the quality of the edited images and show the results in~\tabref{edit}.
As described in~\secref{3_2}, applying the AdaIN optimization causes overfitting that degrades the quality of the edited images.
For instance, applying the AdaIN optimization increases the FID score from $38.68$ to $44.28$ on the face drawing dataset.
One straightforward solution to alleviate this issue is to apply strong weight decay regularizations~(\ie~$w=10^{-2}$).
However, according to the results in 5-th row of~\tabref{recon}, such strong regularizations reduce the reconstruction effectiveness of the AdaIN optimization.
Combining the results in~\tabref{recon} and~\tabref{edit}, we conclude that applying the regularization with the learned hyper-parameter $w$ not only mitigates overfitting but also maintains the efficacy of the AdaIN optimization.
We conduct more analysis of the proposed learning-based regularization in the supplementary materials.

\subsection{Limitations}
The proposed framework has several limitations (see supplemental material for visual examples).
First, since the model learns the multi-stage generation from a training dataset, it fails to produce appealing results if the style of the input image is significantly different from images in the training set.
Second, the uni-modal inference assumption may not be correct.
In practice, the mapping from later stages to previous ones can also be multi-modal.
For instance, the style of the pencil sketches by various artists may be different. 
Finally, artists may not follow a well-staged workflow to create artwork in practice.
However, our main goal is to provide an example workflow to make the artwork creation and editing more feasible, especially for the users who may not be experts in that type of artwork.

\begin{table}[t]
    \footnotesize
	\centering
	\caption{\textbf{Quantitative results of editing.} We use the FID~($\downarrow$) score to evaluate the quality of the edited images $\hat{x}^G_N$ synthesized by the proposed framework. $w$ and LR indicates the hyper-parameter for the weight regularization and applying the learned regularization, respectively.}
	\begin{tabular}{@{}ll ccc@{}} 
	    \toprule
		Optimization & $w$ & Face & Anime & Chair \\
		\midrule
		None & - & $38.68\pm0.44$ & $\mathbf{35.59}\pm0.12$ & $128.4\pm1.50$ \\
		AdaIN & $0$ & $44.28\pm0.45$ & $37.40\pm0.36$ & $97.90\pm1.20$ \\
		AdaIN & $10^{-3}$ & $41.75\pm0.49$ & $38.95\pm0.59$ & $\mathbf{91.68}\pm4.23$\\
		AdaIN & $10^{-2}$ & $\mathbf{38.57}\pm0.94$ & $38.07\pm0.54$ & $99.36\pm7.23$ \\
		\midrule
		AdaIN & LR & $39.40\pm0.21$ & $35.73\pm0.26$ & $95.25\pm0.73$\\
		\bottomrule 
	\end{tabular}
    \vspace{-3mm}
	\label{tab:edit}
\end{table}

\vspace{-1mm}
\section{Conclusions}
\label{sec:conclusions}
\vspace{-1mm}
In this work, we introduce an image generation and editing framework that models the creation stages of an artistic workflow.
We also propose a learning-based regularization for the AdaIN optimization to address the reconstruction problem for enabling non-destructive artwork editing.
Qualitative results on three different datasets show that the proposed framework 1) generates appealing artwork images via multiple creation stages and 2) synthesizes the editing results made by the artists.
Furthermore, the quantitative results validate the effectiveness of the AdaIN optimization and the learning-based regularization.

We believe there are many exciting areas for future research in this direction that could make creating high-quality artwork both more accessible and faster.
We would like to study video sequences of artists as they create artwork to automatically learn meaningful workflow stages that better align with the artistic process.
This could further enable the design of editing tools that more closely align with the operations artists currently perform to iterate on their designs.

\section*{Acknowledgements}
This work is supported in part by the NSF CAREER Grant \#1149783.

\clearpage
%
%
\bibliographystyle{splncs04}
\bibliography{egbib}

\begin{thebibliography}{10}
\providecommand{\url}[1]{\texttt{#1}}
\providecommand{\urlprefix}{URL }
\providecommand{\doi}[1]{https://doi.org/#1}

\bibitem{abdal2019image2stylegan}
Abdal, R., Qin, Y., Wonka, P.: Image2stylegan: How to embed images into the
  stylegan latent space? In: ICCV (2019)

\bibitem{achanta2012slic}
Achanta, R., Shaji, A., Smith, K., Lucchi, A., Fua, P., S{\"u}sstrunk, S.: Slic
  superpixels compared to state-of-the-art superpixel methods. TPAMI
  \textbf{34}(11),  2274--2282 (2012)

\bibitem{lantern}
Adobe: Adobe dimension. \url{https://www.adobe.com/products/dimension.html}
  (2019)

\bibitem{arjovsky2017wasserstein}
Arjovsky, M., Chintala, S., Bottou, L.: Wasserstein gan. In: ICML (2017)

\bibitem{balaji2018metareg}
Balaji, Y., Sankaranarayanan, S., Chellappa, R.: Metareg: Towards domain
  generalization using meta-regularization. In: NIPS (2018)

\bibitem{bau2019semantic}
Bau, D., Strobelt, H., Peebles, W., Wulff, J., Zhou, B., Zhu, J.Y., Torralba,
  A.: Semantic photo manipulation with a generative image prior. ACM TOG (Proc.
  SIGGRAPH)  \textbf{38}(4), ~59 (2019)

\bibitem{brock2018large}
Brock, A., Donahue, J., Simonyan, K.: Large scale gan training for high
  fidelity natural image synthesis. In: ICLR (2019)

\bibitem{chang2015shapenet}
Chang, A.X., Funkhouser, T., Guibas, L., Hanrahan, P., Huang, Q., Li, Z.,
  Savarese, S., Savva, M., Song, S., Su, H., Xiao, J., Yi, L., Yu, F.:
  Shapenet: An information-rich 3d model repository. arXiv preprint
  arXiv:1512.03012  (2015)

\bibitem{chang2018pairedcyclegan}
Chang, H., Lu, J., Yu, F., Finkelstein, A.: Pairedcyclegan: Asymmetric style
  transfer for applying and removing makeup. In: CVPR (2018)

\bibitem{chen2018sketchygan}
Chen, W., Hays, J.: Sketchygan: Towards diverse and realistic sketch to image
  synthesis. In: CVPR (2018)

\bibitem{chen2016infogan}
Chen, X., Duan, Y., Houthooft, R., Schulman, J., Sutskever, I., Abbeel, P.:
  Infogan: Interpretable representation learning by information maximizing
  generative adversarial nets. In: NIPS (2016)

\bibitem{cheng2020rsegvae}
Cheng, Y.C., Lee, H.Y., Sun, M., Yang, M.H.: Controllable image synthesis via
  segvae. In: ECCV (2020)

\bibitem{donahue2019bigbigan}
Donahue, J., Simonyan, K.: Large scale adversarial representation learning. In:
  NIPS (2019)

\bibitem{ghiasi2018dropblock}
Ghiasi, G., Lin, T.Y., Le, Q.V.: Dropblock: A regularization method for
  convolutional networks. In: NIPS (2018)

\bibitem{ghosh2019interactive}
Ghosh, A., Zhang, R., Dokania, P.K., Wang, O., Efros, A.A., Torr, P.H.,
  Shechtman, E.: Interactive sketch \& fill: Multiclass sketch-to-image
  translation. In: CVPR (2019)

\bibitem{goodfellow2014generative}
Goodfellow, I., Pouget-Abadie, J., Mirza, M., Xu, B., Warde-Farley, D., Ozair,
  S., Courville, A., Bengio, Y.: Generative adversarial nets. In: NIPS (2014)

\bibitem{heusel2017gans}
Heusel, M., Ramsauer, H., Unterthiner, T., Nessler, B., Hochreiter, S.: Gans
  trained by a two time-scale update rule converge to a local nash equilibrium.
  In: NIPS (2017)

\bibitem{huang2020semantic}
Huang, H.P., Tseng, H.Y., Lee, H.Y., Huang, J.B.: Semantic view synthesis. In:
  ECCV (2020)

\bibitem{huang2017adain}
Huang, X., Belongie, S.: Arbitrary style transfer in real-time with adaptive
  instance normalization. In: ICCV (2017)

\bibitem{huang2018multimodal}
Huang, X., Liu, M.Y., Belongie, S., Kautz, J.: Multimodal unsupervised
  image-to-image translation. In: ECCV (2018)

\bibitem{hung2018learning}
Hung, W.C., Zhang, J., Shen, X., Lin, Z., Lee, J.Y., Yang, M.H.: Learning to
  blend photos. In: ECCV (2018)

\bibitem{iizuka2016let}
Iizuka, S., Simo-Serra, E., Ishikawa, H.: Let there be color!: joint end-to-end
  learning of global and local image priors for automatic image colorization
  with simultaneous classification. ACM TOG (Proc. SIGGRAPH)  \textbf{35}(4),
  ~110 (2016)

\bibitem{isola2017pix2pix}
Isola, P., Zhu, J.Y., Zhou, T., Efros, A.A.: Image-to-image translation with
  conditional adversarial networks. In: CVPR (2017)

\bibitem{johnson2016perceptual}
Johnson, J., Alahi, A., Fei-Fei, L.: Perceptual losses for real-time style
  transfer and super-resolution. In: ECCV (2016)

\bibitem{karras2017progressive}
Karras, T., Aila, T., Laine, S., Lehtinen, J.: Progressive growing of gans for
  improved quality, stability, and variation. In: ICLR (2018)

\bibitem{karras2019style}
Karras, T., Laine, S., Aila, T.: A style-based generator architecture for
  generative adversarial networks. In: CVPR (2019)

\bibitem{kingma2014adam}
Kingma, D.P., Ba, J.: Adam: A method for stochastic optimization. In: ICLR
  (2015)

\bibitem{krogh1992simple}
Krogh, A., Hertz, J.A.: A simple weight decay can improve generalization. In:
  NIPS (1992)

\bibitem{larsen2015vaegan}
Larsen, A.B.L., S{\o}nderby, S.K., Larochelle, H., Winther, O.: Autoencoding
  beyond pixels using a learned similarity metric. arXiv preprint
  arXiv:1512.09300  (2015)

\bibitem{larsson2016learning}
Larsson, G., Maire, M., Shakhnarovich, G.: Learning representations for
  automatic colorization. In: ECCV (2016)

\bibitem{larsson2016fractalnet}
Larsson, G., Maire, M., Shakhnarovich, G.: Fractalnet: Ultra-deep neural
  networks without residuals. In: ICML (2017)

\bibitem{CelebAMask-HQ}
Lee, C.H., Liu, Z., Wu, L., Luo, P.: Maskgan: Towards diverse and interactive
  facial image manipulation. In: CVPR (2020)

\bibitem{DRIT}
Lee, H.Y., Tseng, H.Y., Huang, J.B., Singh, M.K., Yang, M.H.: Diverse
  image-to-image translation via disentangled representations. In: ECCV (2018)

\bibitem{lee2019drit++}
Lee, H.Y., Tseng, H.Y., Mao, Q., Huang, J.B., Lu, Y.D., Singh, M., Yang, M.H.:
  Drit++: Diverse image-to-image translation via disentangled representations.
  IJCV pp. 1--16 (2020)

\bibitem{lee2019neural}
Lee, H.Y., Yang, W., Jiang, L., Le, M., Essa, I., Gong, H., Yang, M.H.: Neural
  design network: Graphic layout generation with constraints. In: ECCV (2020)

\bibitem{li2019im2pencil}
Li, Y., Fang, C., Hertzmann, A., Shechtman, E., Yang, M.H.: Im2pencil:
  Controllable pencil illustration from photographs. In: CVPR (2019)

\bibitem{li2018closed}
Li, Y., Liu, M.Y., Li, X., Yang, M.H., Kautz, J.: A closed-form solution to
  photorealistic image stylization. In: ECCV (2018)

\bibitem{MSGAN}
Mao, Q., Lee, H.Y., Tseng, H.Y., Ma, S., Yang, M.H.: Mode seeking generative
  adversarial networks for diverse image synthesis. In: CVPR (2019)

\bibitem{nazeri2019edgeconnect}
Nazeri, K., Ng, E., Joseph, T., Qureshi, F., Ebrahimi, M.: Edgeconnect:
  Generative image inpainting with adversarial edge learning. arXiv preprint
  arXiv:1901.00212  (2019)

\bibitem{park2019SPADE}
Park, T., Liu, M.Y., Wang, T.C., Zhu, J.Y.: Semantic image synthesis with
  spatially-adaptive normalization. In: CVPR (2019)

\bibitem{paszke2017pytorch}
Paszke, A., Gross, S., Chintala, S., Chanan, G., Yang, E., DeVito, Z., Lin, Z.,
  Desmaison, A., Antiga, L., Lerer, A.: Automatic differentiation in pytorch.
  In: NIPS workshop (2017)

\bibitem{pathak2016context}
Pathak, D., Krahenbuhl, P., Donahue, J., Darrell, T., Efros, A.A.: Context
  encoders: Feature learning by inpainting. In: CVPR (2016)

\bibitem{portenier2018faceshop}
Portenier, T., Hu, Q., Szabo, A., Bigdeli, S.A., Favaro, P., Zwicker, M.:
  Faceshop: Deep sketch-based face image editing. ACM TOG (Proc. SIGGRAPH)
  \textbf{37}(4), ~99 (2018)

\bibitem{sangkloy2017scribbler}
Sangkloy, P., Lu, J., Fang, C., Yu, F., Hays, J.: Scribbler: Controlling deep
  image synthesis with sketch and color. In: CVPR (2017)

\bibitem{singh2018finegan}
Singh, K.K., Ojha, U., Lee, Y.J.: Finegan: Unsupervised hierarchical
  disentanglement for fine-grained object generation and discovery. In: CVPR
  (2019)

\bibitem{song2019unsupervised}
Song, S., Zhang, W., Liu, J., Mei, T.: Unsupervised person image generation
  with semantic parsing transformation. In: CVPR (2019)

\bibitem{srivastava2014dropout}
Srivastava, N., Hinton, G., Krizhevsky, A., Sutskever, I., Salakhutdinov, R.:
  Dropout: a simple way to prevent neural networks from overfitting. JMLR
  \textbf{15}(1),  1929--1958 (2014)

\bibitem{dropgrad}
Tseng, H.Y., Chen, Y.W., Tsai, Y.H., Liu, S., Lin, Y.Y., Yang, M.H.:
  Regularizing meta-learning via gradient dropout. arXiv preprint
  arXiv:2004.05859  (2020)

\bibitem{crossdomainfewshot}
Tseng, H.Y., Lee, H.Y., Huang, J.B., Yang, M.H.: Cross-domain few-shot
  classification via learned feature-wise transformation. In: ICLR (2020)

\bibitem{tseng2020retrievegan}
Tseng, H.Y., Lee, H.Y., Jiang, L., Yang, W., Yang, M.H.: Retrievegan: Image
  synthesis via differentiable patch retrieval. In: ECCV (2020)

\bibitem{wang2018pix2pixHD}
Wang, T.C., Liu, M.Y., Zhu, J.Y., Tao, A., Kautz, J., Catanzaro, B.:
  High-resolution image synthesis and semantic manipulation with conditional
  gans. In: CVPR (2018)

\bibitem{zhang2018stackgan++}
Zhang, H., Xu, T., Li, H., Zhang, S., Wang, X., Huang, X., Metaxas, D.N.:
  Stackgan++: Realistic image synthesis with stacked generative adversarial
  networks. TPAMI  \textbf{41}(8),  1947--1962 (2018)

\bibitem{zhang2016colorful}
Zhang, R., Isola, P., Efros, A.A.: Colorful image colorization. In: ECCV (2016)

\bibitem{zhang2017real}
Zhang, R., Zhu, J.Y., Isola, P., Geng, X., Lin, A.S., Yu, T., Efros, A.A.:
  Real-time user-guided image colorization with learned deep priors. ACM TOG
  (Proc. SIGGRAPH)  \textbf{9}(4) (2017)

\bibitem{zhu2016generative}
Zhu, J.Y., Kr{\"a}henb{\"u}hl, P., Shechtman, E., Efros, A.A.: Generative
  visual manipulation on the natural image manifold. In: ECCV (2016)

\bibitem{zhu2017unpaired}
Zhu, J.Y., Park, T., Isola, P., Efros, A.A.: Unpaired image-to-image
  translation using cycle-consistent adversarial networks. In: ICCV (2017)

\bibitem{zhu2017toward}
Zhu, J.Y., Zhang, R., Pathak, D., Darrell, T., Efros, A.A., Wang, O.,
  Shechtman, E.: Toward multimodal image-to-image translation. In: NIPS (2017)

\bibitem{zhu2018visual}
Zhu, J.Y., Zhang, Z., Zhang, C., Wu, J., Torralba, A., Tenenbaum, J., Freeman,
  B.: Visual object networks: image generation with disentangled 3d
  representations. In: NIPS (2018)

\end{thebibliography}

\clearpage
\appendix
\section{Appendix}

In this supplementary material, we first present the implementation details for each component of the proposed framework.
Second, we complement the experiment details.
Third, we visualize the learning-based regularization.
Fourth, we show visual examples illustrating the failure cases of the proposed method.
Finally, we present more qualitative results to complement the paper.

\subsection{Implementation Details}
We implement our framework with PyTorch~\cite{paszke2017pytorch}.
The details for each component are described as follows.

\Paragraph{Workflow inference.}
The hyper-parameter $\lambda^1$ in Equation 5 of the paper is assigned to be $10$.
We use the Adam optimizer~\cite{kingma2014adam} with the learning rate of $2\times{10^{-4}}$ and batch size of $8$ for optimizing the model.
We first train each network separately with $450,000$ iterations, then jointly train all the networks in the workflow inference module with $450,000$ iterations.

\Paragraph{Artwork generation.}
We set the hyper-parameter $\lambda^c$ in Equation 3 of the paper to be $1$.
Similar to the training for the workflow inference module, we use the Adam optimizer~\cite{kingma2014adam} with the learning rate of $2\times{10^{-4}}$ and batch size of $8$.
We train each network separately with $1,200,000$ iterations, then jointly train all the networks in the artwork generation module with $600,000$ iterations.
We adopt the objectives in the BicycleGAN~\cite{zhu2017toward} approach for training the artwork generation module, as described in Equation 3 in the paper.
More specifically, the loss $L^\mathrm{bicycle}_i$ in Equation 3 is formulated as
\begin{equation}
\begin{aligned}
    L^\mathrm{bicycle}_i=L^\mathrm{GAN}_i+\lambda^1L^1+\lambda^\mathrm{latent}L^\mathrm{latent}+\lambda^\mathrm{KL}L^\mathrm{KL},
\end{aligned}
\end{equation}
where $L^\mathrm{GAN}_i$ is the hinge version of GAN loss~\cite{brock2018large}, $L^1$ is the $\ell1$ loss between the generated and ground-truth images, $L^\mathrm{latent}$ is the latent regression loss between the predicted and input latent representations, and $L^\mathrm{KL}$ is the KL divergence loss on the latent representations.
Following the setting in the BicycleGAN scheme, we respectively assign the hyper-parameters $\lambda^1$, $\lambda^\mathrm{latent}$, and $\lambda^\mathrm{KL}$ to be $10$, $0.5$, and $0.01$.
We use the network architecture proposed in the MUNIT~\cite{huang2018multimodal} framework (involving AdaIN normalization layers~\cite{huang2017adain}) rather than the U-Net structure in the BicycleGAN framework.

\Paragraph{AdaIN optimization.}
In the editing scenario during the testing phase, we conduct the AdaIN optimization from the first to the last stages sequentially to refine the reconstructed image.
For each stage, we set the hyper-parameters $\lambda^p$, $\alpha$, $T$ in Algorithm 1 in the paper to be $10$, $0.1$ and $150$, respectively.

\Paragraph{Learning-based regularization.}
We summarize the training of the proposed learning-based regularization in Figure 4 of the paper and~\algref{L2R}.
The regularization function is trained separately for each creation stage.
We respectively set the hyper-parameters $\eta$, $T^\mathrm{reg}$, and $\lambda^\mathrm{GAN}$ to be $10^{-3}$, $40000$, and $1$.
We use the Adam optimizer~\cite{kingma2014adam} and the batch size of $1$ for the training.

\begin{algorithm*}[t]
  \caption{Training overview of the learning-based regularization at $i$-th stage}
  \label{alg:L2R}
  \DontPrintSemicolon
  \textbf{Require:} pre-trained generation model $\{E^G_i,G^G_i\}$, learning rate $\eta$, iterations $T^\mathrm{reg}$, importance factor $\lambda^\mathrm{GAN}$\;
  \vspace{1mm}
  $w_i=\mathbf{0.001} \in R^{1\times{8c}}$\;
  \vspace{1mm}
  \While{$t=\{1,\ldots,T^\mathrm{reg}\}$} {
    \vspace{1mm}
    Sample $(x_i, x_{i+1})$ and $x'_i$ from the dataset\;
    \vspace{1mm}
    $z^\mathrm{Ada}_i=E^G_i(x_i)$, $\delta^\mathrm{Ada}_i=\mathbf{0}\in R^{1\times{8c}}$\;
    \vspace{2mm}
    // \textbf{Get reconstructed image before the AdaIN optimization}\;
    $\hat{x}^G_{i+1}=G^G_i(x_i,z^\mathrm{Ada}_i+\delta^\mathrm{Ada}_i)$\;
    \vspace{2mm}
    // \textbf{Optimize incremental term with the regularization function (AdaIN optimization)}\;
    $\Tilde{\delta}^\mathrm{Ada}_i=\delta^\mathrm{Ada}_i-\alpha\left(\bigtriangledown_{\delta^\mathrm{Ada}_i}L_\mathrm{Ada}(\hat{x}^G_{i+1},x_{i+1}) + w_i\delta^\mathrm{Ada}_i\right)$\;
    \vspace{2mm}
    // \textbf{Get the reconstructed image and editing results after the optimization}\;
    $\Tilde{x}^G_{i+1}=G^G_i(x_i,z^\mathrm{Ada}_i+\Tilde{\delta}^\mathrm{Ada}_i)$\;
    $\Tilde{x'}^G_{i+1}=G^G_i(x'_i,z^\mathrm{Ada}_i+\Tilde{\delta}^\mathrm{Ada}_i)$\;
    \vspace{2mm}
    // \textbf{Update the regularization function based on the reconstruction and editing results after the optimization}\;
    $L^\mathrm{L2R}=L^\mathrm{Ada}(\Tilde{x}^G_{i+1},x_{i+1})+\lambda^\mathrm{GAN}L^\mathrm{GAN}(\Tilde{x'}^G_{i+1})$\;
    $w_i=w_i - \eta\bigtriangledown_{w_i}L^\mathrm{L2R}$\;
  }
  \textbf{Return:} $w_i$\;
\end{algorithm*}

\subsection{Experiment Details}
\begin{figure*}[t]
\centering
\includegraphics[width=\linewidth]{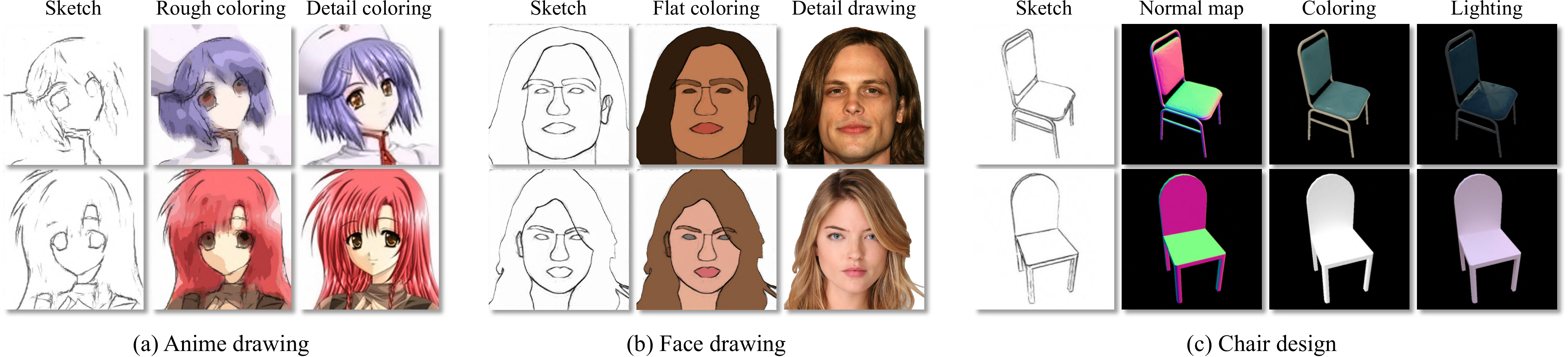}
\vspace{-2mm}
\caption{\textbf{Training examples in each dataset.} For each dataset, we show the example training images at each creation stage.}
\vspace{-3mm}
\label{fig:example}
\end{figure*}

We illustrate how we process each dataset for evaluating the proposed framework.
Example training images in each dataset are shown in~\figref{example}.
In addition, we also describe how we compute FID~\cite{heusel2017gans} score.

\Paragraph{Face drawing dataset.}
We collect the photo-realistic face images from the CelebAMask-HQ dataset~\cite{CelebAMask-HQ}. 
We prepare three design stages for the face drawing dataset: sketch, flat coloring, and detail drawing.
We use the ground-truth attribute segmentation mask to remove the background of the cropped RGB images in the CelebAMask-HQ dataset as the final-stage images.
For the flat coloring, we assign pixels with the median color computed from the corresponding region according to the ground-truth attribute segmentation mask.
Finally, we use the pencil sketch~\cite{li2019im2pencil} model to generate simple sketch images from the flat coloring images.

\Paragraph{Anime drawing dataset.}
We construct the dataset from the anime images in the EdgeConnect~\cite{nazeri2019edgeconnect} dataset.
Three stages are used in this dataset: sketch, rough coloring, detail coloring.
For rough coloring, we first apply the SLIC~\cite{achanta2012slic} super-pixel approach to cluster the pixels in each anime image.
For each cluster, We then compute the median color and assign to the pixels in that cluster.
Finally, we adopt the median filter to smooth the rough coloring images.
As for the sketch, we use the pencil sketch~\cite{li2019im2pencil} scheme to extract the sketch image from the original anime image.

\Paragraph{Chair design.}
We render the chair models in the ShapeNet dataset~\cite{chang2015shapenet} via the photo-realistic renderer~\cite{lantern} for building the dataset.
There are four stages presented in this dataset: sketch, normal map, coloring, and lighting.
We sample two different camera viewpoints for each chair model.
For each viewpoint, we randomly sample from 300 spherical environment maps of diverse indoor and outdoor scenes to render the last-stage image.
For the coloring image, we use a default white lighting environment for the rendering.
We configure the rendering tools to produce the corresponding depth map for each viewpoint and infer the normal map image from the depth map.
Finally, we extract the sketch image from the normal map image using the pencil sketch model~\cite{li2019im2pencil}.

\Paragraph{FID scores.}
We use the official implementation to compute the FID~\cite{heusel2017gans} scores.\footnote{\url{https://github.com/bioinf-jku/TTUR}}
For all experiments, we use the generated images from the whole test set as well as the real images in the training set.
Since we need to re-sample the latent representations for the editing experiments presented in Table 3 in the paper, we conduct $5$ trials for each experiment and report the average results.
We show the FID scores of real images in the test set in \tabref{fid}.
The scores reported in this table can be considered as the lower-bound scores for each task.
\begin{table}[t]
    \footnotesize
	\centering
	\caption{\textbf{FID scores of real images.} We show the FID~($\downarrow$) scores of the real images in the test set to supplement the results in Table 2 and Table 3 of the paper.}
	\vspace{-2mm}
	\begin{tabular}{l ccc} 
	    \toprule
		Datasets & Face & Anime & Chair \\
		\midrule
		Real images & $12.8$ & $16.5$ & $25.3$\\
		\bottomrule 
	\end{tabular}
	\label{tab:fid}
\end{table}
\begin{figure*}[t]
\centering
\includegraphics[width=\linewidth]{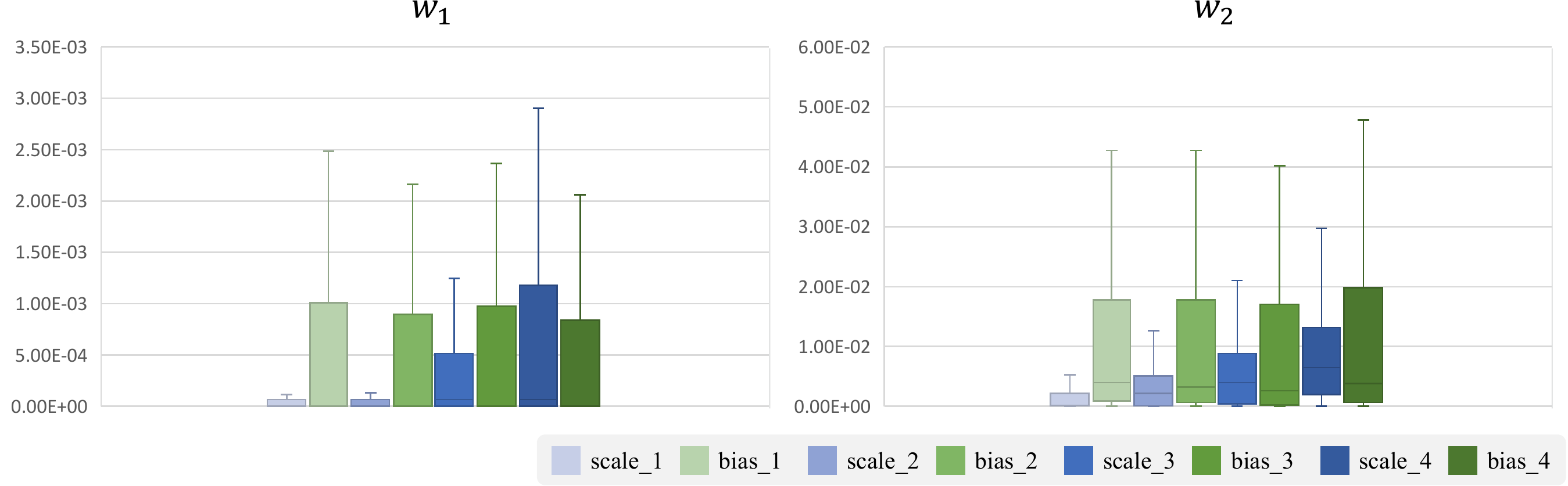}
\vspace{-6mm}
\caption{\textbf{Visualization of the proposed learning-based regularization.} We show the quartile visualization of the hyper-parameter $w_i$ for our learning-based regularization approach trained on the face drawing dataset.
The learned function tends to have stronger regularization on the bias terms.}
\label{fig:vis}
\end{figure*}
\begin{figure}[t]
\centering
\includegraphics[width=0.4\linewidth]{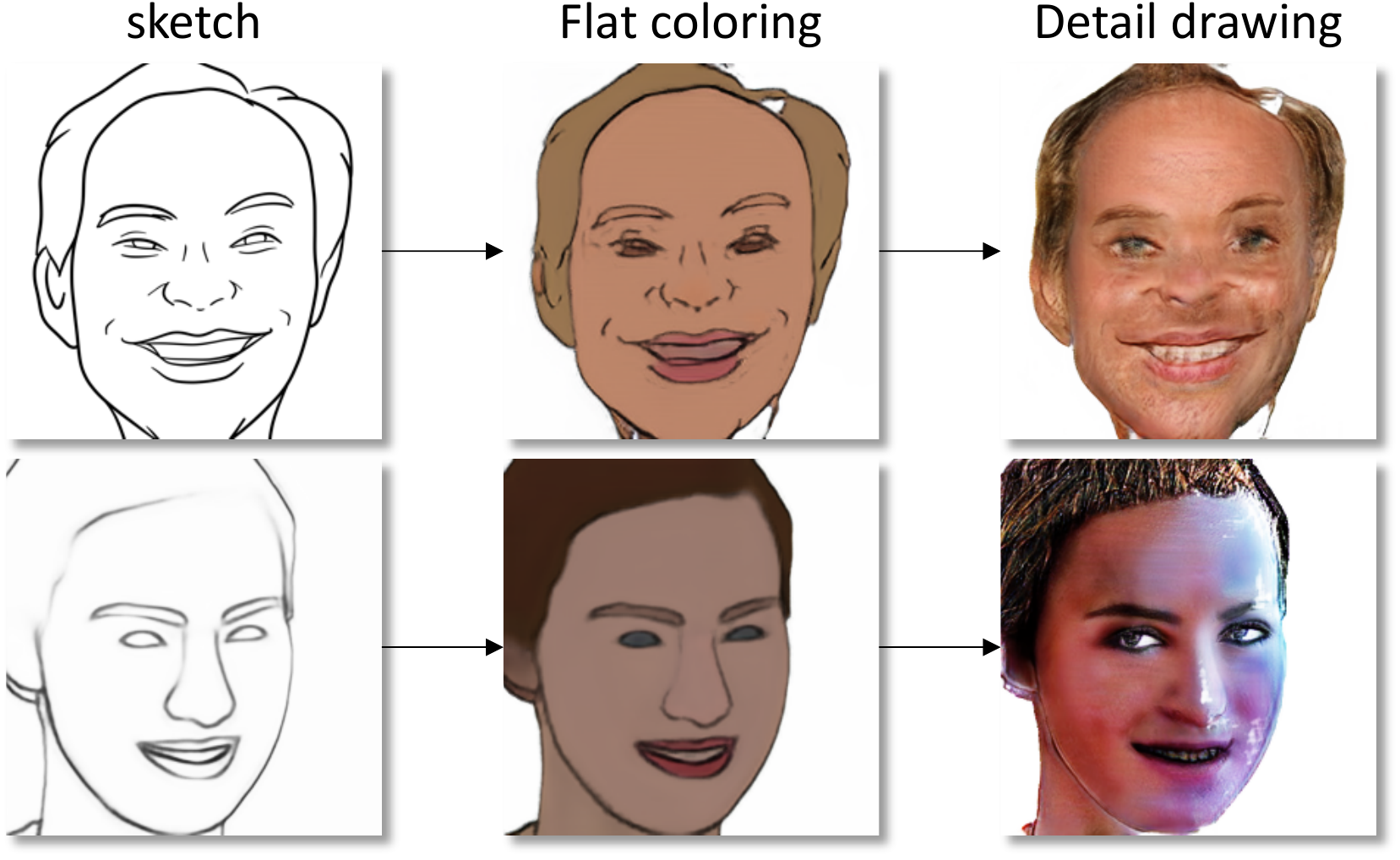}
\vspace{-2mm}
\caption{\textbf{Failure cases.} Our framework fails to generate appealing results (\textit{top}) if the style of the input sketch image is significantly different from that of training images, and (\textit{bottom}) if we use extreme latent representations which are out of the prior distributions we sampled from during the training phase.}
\label{fig:failure}
\vspace{-3mm}
\end{figure}
\begin{figure*}[t]
\centering
\includegraphics[width=\linewidth]{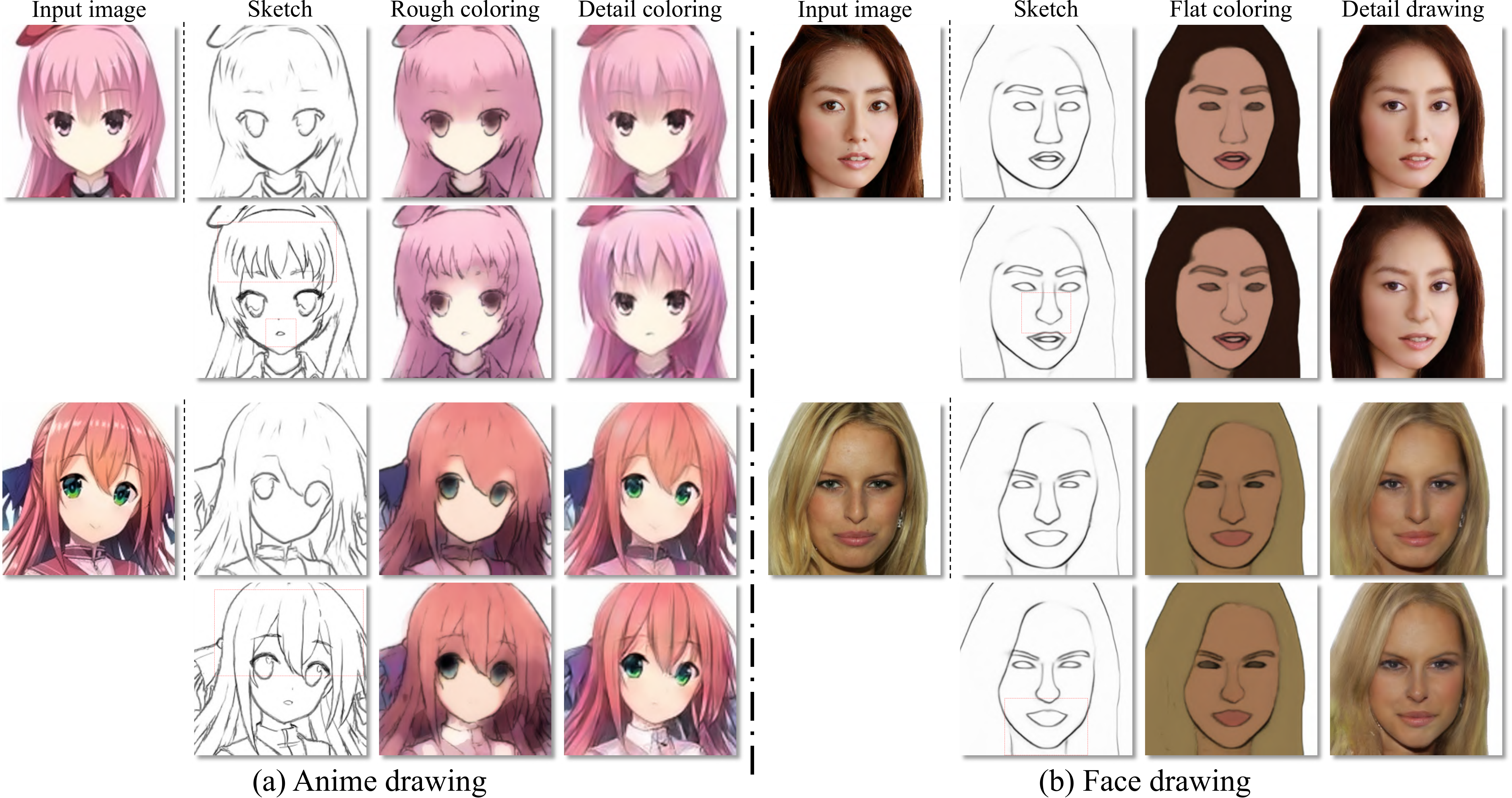}
\vspace{-2mm}
\caption{\textbf{Results of artistic editing.} Given an input artwork image, we ask the artist to edit the inferred sketch image. The synthesis model then produces the corresponding edited artwork. The first row shows the input artwork and inferred images, and the red outlines indicate the edited regions.}
\vspace{-3mm}
\label{fig:artist_supp}
\end{figure*}

\subsection{Additional Experimental Results}

\Paragraph{Visualizing learning-based regularization.}
To understand our learning-based regularization function, we visualize the learned hyper-parameter $w_i$ of the weight decay regularization described in Section 3.3 in the paper and~\algref{L2R}.
The value of the hyper-parameter indicates the strength of the regularization for the AdaIN optimization process.
\figref{vis} shows the visualization of the hyper-parameters trained on the face drawing dataset.
In general, The regularization on the bias terms is stronger than that on the scaling terms.
Since the goal is to minimize the appearance distance between the reconstructed and original input image, the AdaIN optimization tends to complement such discrepancy with the bias terms.
However, as shown in Figure 8 in the paper, such optimization may lead the bias terms to extreme values and make the generation model sensitive to the change~(\ie~editing) of the input image.
The proposed learning-based regularization mitigates the problem by applying stronger regularization on the bias terms, thus encourage the optimization process to modify the scaling terms.
Quantitative results shown in Section 4.3 in the paper validate that the proposed learning-based regularization improves the quality of the editing results.

\Paragraph{Failure cases.}
We observe several failure cases of the proposed framework, which are presented in~\figref{failure}.
First, if the style of the input image is significantly different from the training data, the artwork generation module fails to produce appealing results.
Similar to the Scribbler~\cite{sangkloy2017scribbler} approach, we argue that such a problem may be alleviated by diversifying the style of the training images.
Second, during the creation process, the generation module synthesizes results with artifacts if we use extreme latent representations that are out of the prior distributions we sampled from during the training phase.

\Paragraph{Qualitative results.}
We show more editing results conducted by the artists in~\figref{artist_supp}.
\end{document}